# Testing Implication of Probabilistic Dependencies

S.K.M. Wong
Department of Computer Science
University of Regina
Regina, Saskatchewan
Canada, S4S 0A2
wong@cs.uregina.ca

## Abstract

Axiomatization has been widely used for testing logical implications. This paper suggests a non-axiomatic method, the *chase*, to test if a new dependency follows from a given set of probabilistic dependencies. Although the chase computation may require exponential time in some cases, this technique is a powerful tool for establishing nontrivial theoretical results. More importantly, this approach provides valuable insight into the intriguing connection between relational databases and probabilistic reasoning systems.

## 1 INTRODUCTION

In probability theory, the notion of dependencies (independencies) play an important role. The knowledge of conditional independencies, in particular, is essential for developing a viable probabilistic reasoning system [9, 12, 13].

Given a set of probabilistic dependencies, there are additional dependencies *implied* by this set in the sense that any joint probability distribution that satisfies the original set must also satisfy the additional dependencies. Developing a qualitative method for testing logical implication of dependencies is important for many reasons. First, it enables us to derive interesting and powerful theorems that may or may not be obvious from the numerical representation of probabilities. Second, in the design of a probabilistic inference system, we often need to know whether one dependency is implied by a given set of dependencies.

*Axiomatization* has been widely used for determining logical implications [2, 14, 15, 16, 17, 19]. In this approach, a finite set of complete inference rules are introduced for a particular class of dependencies. These rules are used to generate symbolic proofs for new dependencies in a manner analogous to proofs in mathematical logic. In this paper, we adopt an alternative method from relational database theory [7, 11] for testing logical implication of probabilistic dependencies. We use *tableaux* and an operation on tableaux, the *chase*, to test if a new dependency follows from the initial set of dependencies. Our study will focus on the *generalized acyclic join dependency* (GAJD) [20, 21]. Probabilistic conditional independencies are a subclass of this dependency.

The chase computation may require exponential time in some cases [11]. However, this approach provides valuable insight into the intriguing connection between relational database and probabilistic reasoning systems. On the practical side, the chase technique is a powerful tool for establishing some important theoretical results. For example, based on this technique, one can show that a GAJD is equivalent to a set of probabilistic conditional independencies. The chase method can also be used to study the optimization problems in probabilistic reasoning. (The results of these studies will be reported in a separate paper.)

This paper is organized as follows. In Section 2, we first establish the fact that probabilistic knowledge can be represented as a generalized relational database. In particular, we show that a decomposable joint probability distribution can be conveniently represented as a GAJD. Within this framework, one can view evidential reasoning simply as processing a conjunctive query in a generalized relational database system. In Section 3, we develop the chase algorithm. We show in Section 4 how the chase technique is applied to testing dependency implications. The conclusion is presented in Section 5.

## 2 GENERALIZED RELATIONAL DATABASE

It has been pointed out that there exists an intriguing connection between relational database and prob-abilistic reasoning systems [20, 21]. In this section, we first briefly describe some database concepts pertinent to our discussion. Then we introduce our extended relational data model. We show that a decomposable joint probability distribution is equivalent to a generalized acyclic join dependency (GAJD) in this model, and probabilistic conditional independence is a special case of GAJD. Representation of a GAJD by a *tableau* will be discussed in Section 3.

## 2.1 RELATIONAL DATABASE CONCEPTS

Let $\mathcal{N}$ be a finite set of variables called attributes. We will use upper case letters $A, B, C, ...$ to denote a single attribute and $..., X, Y, Z$ to represent a subset of attributes. Each attribute $A \in \mathcal{N}$ takes on values from a domain $V_A$. Consider a subset of attributes $X = \{A_1, A_2, ..., A_l\} \subseteq \mathcal{N}$. Let $V_X = V_{A_1} \cup V_{A_2} \cup ... \cup V_{A_l}$ be the domain of $X$. A X-tuple $t_X$ is a mapping from $X$ to $V_X$, i.e., $t_X : X \longrightarrow V_X$, with the restriction that for each attribute $A \in X$, $t_X[A]$ must be in $V_A$. (We write $t$ instead of $t_X$ if $X$ is understood.) Thus $t$ is a mapping that associates a value in $V_X$ with each corresponding attribute in $X$. If $Y$ is a subset of $X$ and $t$ a X-tuple, then $t[Y]$ denotes the Y-tuple obtained by restricting the mapping to $Y$. Let $y = t[Y]$. We call $y$ a Y-value which is also referred to as a *configuration* of $Y$. A X-relation $r$ (or a relation r over $X$, or simply a relation $r$ if $X$ is understood) is a finite set of X-tuples or X-values. If $r$ is a X-relation and $Y$ is a subset of $X$, then by $r[Y]$, the projection of relation $r$ onto $Y$, we mean the set of tuples $t[Y]$, where $t$ is in $r$.

We define a database scheme $\mathbf{R} = \{R_1, R_2, ..., R_N\}$ to be a set of subsets of $\mathcal{N}$. We call the $R_i$'s relation schemes. If $r_1, r_2, ..., r_N$ are relations, where $r_i$ is a relation over $R_i$ ($1 \leq i \leq N$), then we call $\mathbf{r} = \{r_1, r_2, ..., r_N\}$ a database over $\mathbf{R}$. The join (natural join) of the relations in $\mathbf{r}$ (where the join is denoted by either $r_1 \bowtie r_2 \bowtie ... \bowtie r_N$ or $\bowtie \mathbf{r}$) is the set of all tuples $t$ with attributes $R_1 \cup R_2 \cup ... \cup R_N$, such that $t[R_i]$ is in $r_i$ for each $i$ ($1 \leq i \leq N$). We say that a relation $r$ with attributes $R = R_1 \cup R_2 \cup ... \cup R_N$ obeys the *join dependency* $\bowtie \{R_1 \cup R_2 \cup ... \cup R_N\} = \bowtie \mathbf{R}$, if $r = \bowtie\{r_1, r_2, ..., r_N\}$, where $r_i = r[R_i]$, for $1 \leq i \leq N$. It follows that the join dependency $\bowtie\{R_1 \cup R_2 \cup ... \cup R_N\}$ holds for a relation $r$ if and only if $r$ contains each tuple $t$ for which there are tuples $t_1, t_2, ..., t_N$ of $r$ (not necessarily distinct) such that $t_i[R_i] = t[R_i]$ for each $i$ ($1 \leq i \leq N$).

*Multivalued dependency* (MVD) [6, 7, 8] is a special case of join dependency (JD) [3, 11]. We say that the MVD $X \twoheadrightarrow Y$ holds for relation $r$ if for any $t_1, t_2 \in r$ with $t_1[X] = t_2[X]$, there exists a tuple $t_3 \in r$ such that $t_3[X] = t_1[X]$, $t_3[Y] = t_1[Y]$ and $t_3[R - XY] = t_2[R - XY]$. By XY, we mean $X \cup Y$.

## 2.2 AN EXTENDED RELATIONAL DATA MODEL

In the proposed relational data model [20, 21], each relation $\Phi_R$ represents a real-valued function $\phi_R$ on a set of attributes $R = \{A_1, A_2, ..., A_l\}$ as shown in Figure 1, where $t_{ij} \in V_{A_j}$, i.e., $t_i = (t_{i1}, t_{i2}, ..., t_{il}) \in V_R$ is a configuration (tuple) of $R$. The function $\phi_R(t_i)$, defines the values of the attribute $f_{\phi_R}$ in relation $\Phi_R$. The semantic interpretation of the function $\phi_R$ would depend very much on the particular application.

In the conventional database model, for example, $\phi(t)$ could be interpreted as the number of tuples $t$ in a relation, if one is interested in keeping track of duplicate tuples resulting from a projection. Let $\Phi_U$ denote a *universal* relation with $\phi(t) = 1$ for all tuples of $U = A_1, A_2, ..., A_l$. The relation $\Phi_R$ shown in Figure 1 can be interpreted as the relation obtained by projecting $\Phi_U$ onto $R = A_1 A_2 ... A_l \subseteq U$, where $\phi(t_i)$ is the number of tuples with $t_i = t[R]$ in the original relation. Clearly, it is not necessary to use such a function $\phi$ to define a relation, if counting of duplicate tuples is not an issue. We will show that the conventional relational database model is indeed a special case of the extended data model introduced here.

On the other hand, in a probabilistic model, for example, the relation $\Phi_R$ shown in Figure 1 represents a marginal probabilistic distribution. That is, the function $\phi_R(t)$ on $R$, which defines the values of the attribute $f_{\phi_R}$ in relation $\Phi_R$, is a joint distribution (a marginal distribution).

$$\Phi_R = \begin{array}{|ccccc|} \hline A_1 & A_2 & \ldots\ldots & A_l & f_{\phi_R} \\ \hline t_{11} & t_{12} & \ldots\ldots & t_{1l} & \phi_R(t_1) \\ t_{21} & t_{22} & \ldots\ldots & t_{2l} & \phi_R(t_2) \\ . & . & . & . & . \\ . & . & . & . & . \\ . & . & . & . & . \\ t_{s1} & t_{s2} & \ldots\ldots & t_{sl} & \phi_R(t_s) \\ \hline \end{array}$$

Figure 1: The relation $\Phi_R$ representing a function $\phi_R$ on $R = \{A_1, A_2, ..., A_l\}$.

We can define an *inverse* relation $(\Phi_R)^{-1}$ for $\Phi_R$, by setting $\phi_R(t_i)^{-1} = \frac{1}{\phi_R(t_i)}$, for each $t_i$ ($1 \leq i \leq s$) with $\phi_R(t_i) \neq 0$. The reason for introducing such an inverse relation will become clear when a specific application is considered.

Apart from the *select, project* and *natural join* operators in a standard relational system, we define here two new relational operators called *marginalization* and *product join*.

1. *Marginalization*

    Let $X$ be a subset of attributes of $R$. The operator of marginalization is denoted by the symbol $\downarrow$. The marginal $\phi_R^{\downarrow X}$ of $\Phi_R$ is a relation on $X \cup \{f_{\phi_R}\}$. We can construct $\Phi_R^{\downarrow X}$ from $\Phi_R$ as follows:

    (a) Project the relation $\Phi_R$ on the set of attributes $X \cup \{f_{\phi_R}\}$, without eliminating identical configurations (tuples).

    (b) Let $t$ be a tuple in $\Phi_R[R]$. For every configuration $t_X = t[X]$, replace the set of configurations of $X \cup \{f_{\phi_R}\}$ in the relation obtained from step (a) by the *singleton* configuration:

    $$t_X * (\sum_{t_{R-X}} \phi_R(t_X * t_{R-X})),$$

    where $t_{R-X} = t[R - X]$ and $t = t_X * t_{R-X}$. The symbol $*$ denotes concatenation of two tuples.

2. *Product Join*

    Consider two relations $\Phi_X, \Psi_Y$ defined respectively by functions $\phi_X$ and $\psi_Y$. The *product join* of $\Phi_X$ and $\Psi_Y$, written $\Phi_X \times \Psi_Y$, is defined as follows:

    (i) Compute the natural join, $\Phi_X \bowtie \Psi_Y$, of relations $\Phi_X$ and $\Psi_Y$.

    (ii) Add a new column labeled by the attribute $f_{\phi_X \cdot \psi_Y}$ to the relation $\Phi_X \bowtie \Psi_Y$. The values of the attribute $f_{\phi_X \cdot \psi_Y}$ are defined by the product $\phi_X(t[X]) \cdot \psi_Y(t[Y])$, where $t$ is a configuration of $XY$ such that $t[X] = t_X \in \Phi_X[X]$ and $t[Y] = t_Y \in \Psi_Y[Y]$.

    (iii) The resultant relation $\Phi_X \times \Psi_Y$ is obtained by projecting the relation obtained from step (ii) on the set of attributes $XY \cup \{f_{\phi_X \cdot \psi_Y}\}$.

Examples illustrating the marginalization and product join operators are given in [21]. It should be noted that both the marginalization and product join operators introduced above can be defined more generally. For example, in step 2(ii), the values of the attribute $f_{\phi_X \cdot \psi_Y}$ can be defined as $\phi_X(t[X]) \circ \psi_Y(t[Y])$, where $\circ$ is a binary operator (not necessarily the ordinary multiplication operator). As in this paper we focus on the study of the relationship between relational and probabilistic systems, we have deliberately chosen ordinary multiplication to define the *product* of $\phi_X$ and $\psi_Y$. In this context, we have of course in mind the notion of probabilistic conditional independence, and a marginal relation would represent a marginal distribution. It is understood that in general the choice of the product operator in step 2(ii) and 1(b) of marginalization depends on the specific problem being modeled.

It is perhaps worth mentioning at this point that given any relation $\Phi_R$, the product join, $\Phi_R \times (\Phi_R)^{-1}$, is a *unit* relation, i.e., $\phi_R(t) \cdot \phi_R^{-1}(t) = 1$ for all $t$'s in $\Phi_R[R]$. In fact, inverse relations become quite useful in the discussion of the probabilistic model.

### 2.3 GENERALIZED ACYCLIC JOIN DEPENDENCY (GAJD)

First, let us introduce some notions of graph theory pertinent to our discussion. A hypergraph is a pair $(\mathcal{N}, \mathbf{R})$, where $\mathcal{N}$ is a finite set of nodes (attributes) and $\mathbf{R}$ is a set of edges (hyperedges) which are arbitrary subsets of $\mathcal{N}$ [4, 18]. If the nodes are understood, we will use $\mathbf{R}$ to denote the hypergraph $(\mathcal{N}, \mathbf{R})$. An ordinary undirected graph (without self-loops) is, of course, a hypergraph whose every edge is of size two. We say an element $R_i$ in a hypergraph $\mathbf{R}$ is a *twig* if there exists another element $R_j$ in $\mathbf{R}$, distinct from $R_i$, such that $(\cup(\mathbf{R} - \{R_i\})) \cap R_i = R_i \cap R_j$. We call any such $Rj$ a *branch* for the twig $R_i$. A hypergraph $\mathbf{R}$ is a *hypertree* [10, 18] if its elements can be ordered, say $R_1, R_2, ..., R_N$, so that $R_i$ is a twig in $\{R_1, R_2, ..., R_i\}$, for $i = 2, ..., N$. We call any such ordering a *tree (hypertree) construction ordering* for $\mathbf{R}$. Given a tree construction ordering $R_1, R_2, \ldots, R_N$, we can choose, for $i$ from 2 to $N$, an integer $j(i)$ such that $1 \leq j(i) \leq i - 1$ and $S_{j(i)}$ is a branch for $R_i$ in $\{R_1, R_2, \ldots, R_i\}$. We call a function $j(i)$ that satisfies this condition a *branching* for $\mathbf{R}$ and $R_1, R_2, \ldots, R_N$. For example, let $\mathcal{N} = \{A_1, A_2, ..., A_6\}$. Consider a hypergraph $\mathbf{R} = \{R_1 = \{A_1, A_2, A_3\}, R_2 = \{A_1, A_2, A_4\}, R_3 = \{A_2, A_3, A_5\}, R_4 = \{A_5, A_6\}\}$. This hypergraph is a hypertree, as there exists a tree construction ordering, $R_3, R_1, R_2, R_4$. Furthermore, the branching function for this ordering is $j(1) = 3$, $j(2) = 1$, $j(4) = 3$.

Given a tree construction ordering $R_1, R_2, \ldots, R_N$ for a hypertree $\mathbf{R}$ and a branching function $j(i)$ for this ordering, we can construct the following set of subsets: $\mathcal{L} = \{R_{j(2)} \cap R_2, R_{j(3)} \cap R_3, ..., R_{j(N)} \cap R_N\}$. It is important to note that this set $\mathcal{L}$ is independent of the tree construction ordering, i.e., $\mathcal{L}$ is the same for any tree construction ordering of a given hypertree. We call $\mathcal{L}$ the *interaction set* of the hyperedges in $\mathbf{R}$.

Consider a relation $\Psi_R$ over the set of attributes $S = R \cup \{f_{\Psi_R}\} = R_1 \cup R_2 \cup \ldots \cup R_N \cup \{f_{\Psi_R}\}$, where $\Psi_R$ represents a joint probability distribution over the variables $R = R_1 \cup R_2 \cup \ldots \cup R_N$. Suppose the hypergraph $\mathbf{R} = \{R_1, R_2, \ldots, R_N\}$ is a hypertree. We say that $\Psi_R$ satisfies the *generalized acyclic join de-* *pendency* (GAJD), written $\bigotimes \mathbf{R}[\Psi_R]$, if

$$\Psi_R = (\ldots((\Psi_R^{\downarrow R_1} \otimes \Psi_R^{\downarrow R_2}) \otimes \Psi_R^{\downarrow R_3})\ldots \otimes \Psi_R^{\downarrow R_N}),$$

which is a *sequential monotone join expression*. The monotone join operator $\otimes$ is defined by: for any $R_i, R_j \subseteq R$,

$$\Psi_R^{\downarrow R_i} \otimes \Psi_R^{\downarrow R_j} = \Psi_R^{\downarrow R_i} \times \Psi_R^{\downarrow R_j} \times (\Psi_R^{\downarrow R_i \cap R_j})^{-1},$$

where $\times$ is the product join operator and $(\Psi_R^{\downarrow R_i \cap R_j})^{-1}$ denotes the inverse relation $\Psi_R^{\downarrow R_i \cap R_j}$. It should be noted that the sequence, $R_1, R_2, \ldots, R_N$, is a hypertree construction ordering of the hypergraph $\mathbf{R}$.

It should be noted that probabilistic conditional independence is a special case of GAJD. This can be easily seen as follows. Let $\mathbf{R} = \{R_1, R_2\}$. In this case, the hypergraph $\mathbf{R}$ is always a hypertree. Let $\Psi_\mathbf{R}$ be defined by a probability distribution $\psi_R$. Clearly, the condition,

$$\begin{aligned}\Psi_R &= m_\mathbf{R}(\Psi_R) = \Psi_R^{\downarrow R_1} \otimes \Psi_R^{\downarrow R_2} \\ &= (\Psi_R^{\downarrow R_1}) \times \Psi_R^{\downarrow R_2} \times (\Psi_R^{\downarrow R_1 \cap R_2})^{-1},\end{aligned}$$

can be equivalently expressed as:

$$\psi_R = \frac{\psi_R(R_1) \cdot \psi_R(R_2)}{\psi_R(R_1 \cap R_2)},$$

namely, $R_1$ and $R_2$ are conditionally independent given $R_1 \cap R_2$.

### 2.4 REPRESENTATION OF A DECOMPOSABLE PROBABILITY DISTRIBUTION AS A GAJD

By the chain rule of probability, any joint distribution $\phi(a_1 a_2 \ldots a_l)$ can be expressed as:

$$\phi(a_1 a_2 \ldots a_l) = \phi(a_1) \cdot \phi(a_2|a_1) \cdot \ldots \cdot \phi(a_l|a_1 a_2 \ldots a_{l-1}),$$

where $a_i \in V_{A_i}$, i.e., $a_i$ is an $A_i$-value of attribute $A_i \in R = \{A_1, A_2, \ldots, A_l\}$. For convenience, we have written $\phi_R(a_1, a_2, \ldots, a_l)$ as $\phi(a_1 a_2 \ldots a_l)$. The above identity is particularly useful in using conditional independencies to simplify the representation of a joint distribution. Consider, for example, a distribution on the set of variables $\{A_1, A_2, A_3, A_4, A_5, A_6\}$:

$$\begin{aligned}\phi(a_1 a_2 a_3 a_4 a_5 a_6) &= \phi(a_1) \cdot \phi(a_2|a_1) \cdot \\ &\quad \phi(a_3|a_1 a_2) \cdot \phi(a_4|a_1 a_2 a_3) \cdot \\ &\quad \phi(a_5|a_1 a_2 a_3 a_4) \cdot \phi(a_6|a_1 a_2 a_3 a_4 a_5).\end{aligned}$$

Note that for convenience, $\phi(a_1, a_2, a_3, \ldots)$ is written as $\phi(a_1 a_2 a_3 \ldots)$. Suppose the following conditional independencies hold:

$\phi(a_3|a_1 a_2) = \phi(a_3|a_1)$,
$\phi(a_4|a_1 a_2 a_3) = \phi(a_4|a_1 a_2)$,
$\phi(a_5|a_1 a_2 a_3 a_4) = \phi(a_5|a_2 a_3)$,
$\phi(a_6|a_1 a_2 a_3 a_4 a_5) = \phi(a_6|a_5)$.

The above joint distribution $\phi(a_1 a_2 a_3 a_4 a_5 a_6)$ can then be simplified to:

$$\begin{aligned}\phi(a_1 a_2 a_3 a_4 a_5 a_6) &= \phi(a_1) \cdot \phi(a_2|a_1) \cdot \phi(a_3|a_1) \cdot \\ &\quad \phi(a_4|a_1 a_2) \cdot \phi(a_5|a_2 a_3) \cdot \\ &\quad \phi(a_6|a_5).\end{aligned}$$

In fact, this distribution can be depicted by a directed acyclic graph (DAG) which is known as a *Bayesian network* [14].

For certain applications, it is much more convenient to represent a joint distribution by a chordal (triangulated) undirected graph. A determined undirected graph G depicts the following distribution:

$$\phi(a_1 a_2 a_3 a_4 a_5 a_6) = \frac{\phi(a_1 a_2 a_3) \cdot \phi(a_1 a_2 a_4) \cdot \phi(a_2 a_3 a_5) \cdot \phi(a_5 a_6)}{\phi(a_1 a_2) \cdot \phi(a_2 a_3) \cdot \phi(a_5)}.$$

We say that the above distribution is *decomposable* [14] (relative to the graph G). Note that each maximal clique in G represents a marginal distribution in the numerator of the above equation.

Consider a chordal undirected graph G representing a joint probability distribution $\phi_R$, i.e., $\phi_R$ is decomposable relative to G. Let $\mathbf{R}(G)$ be the hypergraph whose hyperedges are precisely the maximal cliques of G. Thus, $\mathbf{R}(G)$ is both chordal and *conformal* [3, 4], namely, $\mathbf{R}(G)$ is a hypertree. Let $\mathcal{L}$ denote the interaction set of the hyperedges in $\mathbf{R}(G)$ as defined in Section 3.2.

**Lemma 1.** [9, 14]. If a joint probability distribution $\phi$ is decomposable relative to a chordal undirected graph G, then $\phi$ can be written as a product of the marginal distributions of the maximal cliques of G divided by a product of the marginal distributions of the interaction set of $\mathbf{R}(G)$.

It should perhaps be noted that the computation of marginal distributions is a major problem in practical applications of Bayesian networks as it may easily become intractable [5]. Fortunately, many efficient algorithms based on the techniques of local propagation [10, 18] have been developed for computing the marginals of a factorized joint probability distribution.

Suppose a joint distribution $\phi_R$ is decomposable relative to a chordal graph G. Let $\mathbf{R} = \{R_1, R_2, \ldots, R_N\}$ denote the set of hyperedges of the hypertree $\mathbf{R}(G)$. Let $R = R_1 \cup R_2 \cup \ldots \cup R_N$. Each hyperedge $R_i$ in $\mathbf{R}(G)$ defines a marginal distribution $\phi_i$ of $\phi_R$. Let $\mathcal{L} = \{R_{j(2)} \cap R_2, R_{j(3)} \cap R_3, \ldots, R_{j(N)} \cap R_N\}$ be the intersection set of $\mathbf{R}(G)$, in which we have tacitly assumed that the sequence $R_1, R_2, \ldots, R_N$ is a tree construction ordering for $\mathbf{R}(G)$. The joint probability distribution $\phi_R$ can be represented as a relation $\Phi_R$ over the set of attributes $S = R \cup \{f_{\phi_R}\}$, where the values of the attribute $f_{\phi_R}$ are defined by the function $\phi_R$. Similarly, each marginal distribution $\phi_{R_i}$ ($1 \leq i \leq N$) is represented by a relation $\Phi_R^{\downarrow R_i}$ over $S_i = R_i \cup \{f_{\phi_i}\}$, where the values of the attribute $f_{\phi_{R_i}}$ are defined by the function $\phi_{R_i}$.

By Lemma 1 and the definition of product join, the relation $\Phi_R$ over $S$ can be expressed as:

$$\Phi_R = \Phi_R^{\downarrow R_1} \times ... \times \Phi_R^{\downarrow R_N} \times (\Phi_R^{\downarrow R_{j(2)} \cap R_2})^{-1}$$
$$\times ... \times (\Phi_R^{\downarrow R_{j(N)} \cap R_N})^{-1}. \quad (2)$$

Since the sequence $R_1, R_2, ..., R_N$ is a tree construction ordering for $\mathbf{R}(G)$, we have for $1 \leq j(i) \leq i-1$ and $i = 2, 3, ..., N$:

$$(R_1 \cup R_2 \cup ... \cup R_{i-1}) \cap R_i = R_{j(i)} \cap R_i,$$

where $R_{j(i)}$ is a branch of the twig $R_i$ in the hypertree. Thus Equation 2 can be written as a sequential monotone join expression as defined in Section 2.3:

$$\Phi_R = (....((\Phi_R^{\downarrow R_1} \otimes \Phi_R^{\downarrow R_2}) \otimes \Phi_R^{\downarrow R_3}).... \otimes \Phi_R^{\downarrow R_N})$$
$$= m_\Phi(\Phi_R).$$

This means that the relation $\Phi_R$ satisfies the GAJD $\otimes \mathbf{R}[\Phi_R]$.

**Theorem 1** *A decomposable joint probability distribution is equivalent to a generalized acyclic join dependency.*

## 3  MARGINALIZE-PRODUCT-JOIN MAPPINGS, TABLEAUX, AND THE CHASE

As mentioned in the introduction, the main objective of this paper is to suggest a procedure, called the *chase*, for testing logical implications of probabilistic dependencies (independencies), the generalized acyclic join dependencies in particular. This method provides an alternative approach to using axiomatization [14, 17, 19] for inferring new dependencies from a given set of dependencies.

### 3.1  MARGINALIZE-PRODUCT-JOIN MAPPINGS

Consider a decomposable joint probability distribution $\phi_R$ on the set of variables $R = \{A_1, A_2, ..., A_m\}$, and a hypertree $\mathbf{R} = \{R_1, R_2, ..., R_N\}$ with $R = R_1 \cup R_2 \cup ... \cup R_N$. The probability distribution $\phi_R$ can be represented as a relation $\Phi_R$ (see Section 2.4) over the set of attributes $S = R \cup \{f_{\phi_R}\}$, where the values of the attribute $f_{\phi_R}$ are defined by the function $\phi_R$.

Likewise, each marginal distribution $\phi_i$ of $\phi$ on $R_i$ ($1 \leq i \leq N$) is represented by a relation $\Phi_R^{\downarrow R_i}$ over $S_i = R_i \cup \{f_{\phi_i}\}$. The *marginalize-product-join mapping*, written $m_\mathbf{R}(\Phi_R)$, is a function on relations over $S$ defined by:

$$m_\mathbf{R}(\Phi_R) = (...((\Phi_R^{\downarrow R_1} \otimes \Phi_R^{\downarrow R_2}) \otimes \Phi_R^{\downarrow R_3})... \otimes \Phi_R^{\downarrow R_N}),$$

i.e., $m_\mathbf{R}(\Phi_R)$ is a sequential monotone join expression as defined in Section 2.3. Saying that a relation $\Phi_R$ (representing a probability distribution $\phi_R$) satisfies the GAJD, $\otimes \mathbf{R}[\Phi_R]$, is the same as saying $m_\mathbf{R}(\Phi_R) = \Phi_R$.

Very often, we are not interested in all possible relations on $S$. We are primarily interested in some subset, say $\mathbf{P}$. As $\mathbf{P}$ may be an infinite set, it cannot be described by enumeration. Instead, it can be described by a set of *constraints* (such as GAJDs). Let $\mathbf{C}$ denote a set of constraints, and let $\mathbf{P} = SAT_S(\mathbf{C})$ denote the set of relations that satisfy all the constraints in $\mathbf{C}$. We can now precisely define the notion of *logical implication* as follows. Let $c$ denote a single constraint. We say that $\mathbf{C}$ logically implies $c$, written $\mathbf{C} \models c$, if $SAT(\mathbf{C}) \subseteq SAT(c)$. (Note that we drop the subscript $S$ in $SAT_S$ if no confusion arises.) In subsequent sections, we will develop a procedure to test if a given set of constraints $\mathbf{C}$ logically implies a GAJD, say $\otimes \mathbf{R}[\Phi_R]$, namely, we want to test if $\mathbf{C} \models \otimes \mathbf{R}[\Phi_R]$ holds.

### 3.2  TABLEAUX AS MAPPINGS

Similar to relational databases [1, 11], this section presents a tabular method for representing marginalize-product-join mappings. A *tableau* is similar to a relation $\Phi_R$ in the extended data model, except, in places of *values*, a tableau is defined by a set of *variables*. Consider for example, the following tableau $T$ with $S = R \cup \{f_{\phi_R}\} = \{A_1, A_2, A_3, A_4\} \cup \{f_{\phi_R}\}$:

|     | $A_1$ | $A_2$ | $A_3$ | $A_4$ | $f_{\phi_R}$ |
|-----|-------|-------|-------|-------|--------------|
|     | $a_1$ | $b_1$ | $a_3$ | $b_2$ | $p_1 = \phi_R(a_1, b_1, a_3, b_2)$ |
| $T =$ | $b_3$ | $a_2$ | $a_3$ | $b_4$ | $p_2 = \phi_R(b_3, a_2, a_3, b_4)$ |
|     | $a_1$ | $b_5$ | $a_3$ | $a_4$ | $p_3 = \phi_R(a_1, b_5, a_3, a_4)$ |

The set $S$ of attributes labels the columns in the tableau; $S$ is referred to as the scheme of the tableau. The $p$'s are the variables of the attribute $f_{\phi_R}$. The subscripted $a$'s are called *distinguished* variables, and the subscripted $b$'s are called *nondistinguished* variables. Each variable may appear in only one column. Furthermore, only one distinguished variable may appear in each column. By convention, the distinguished variable $a_i$ will be the one that appears in the column of the attribute $A_i$. We assume in this paper that every distinguished variable appears at least once.

Let $T$ be a tableau and let

$$V = \{a_1, a_2, \ldots, a_l, b_1, b_2, \ldots, p_1, p_2, \ldots\}$$

denote the set of its variables. A *valuation* $\delta$ for $T$ is a mapping from $V$ to $V_S = V_{A_1} \times V_{A_2} \times \ldots \times V_{A_l} \times V_{A_{l+1}}$ such that $\delta(v)$ is in $V_{A_i}$ if $v$ is in the column of $A_i$, where $S = \{A_1, A_2, \ldots, A_l, f_{\phi_R}\}$, $V_{A_{l+1}} = V_{f_{\phi_R}}$, and $V_{A_i}$ is the domain of $A_i$. We extend valuations to apply to rows (tuples) of $T$ in the obvious manner: if $\mathbf{w}$ is the row $< v_1, v_2, \ldots, v_{l+1} >$ in $T$, then $\delta(\mathbf{w})$ is the row $< \delta(v_1), \delta(v_2), \ldots, \delta(v_{l+1}) >$, where $\delta(v_{l+1}) = \phi_R(\delta(\mathbf{w}[R]))$. Applying $\delta$ to the entire tableau, we have:

$$\delta(T) = \{\delta(\mathbf{w}) \mid \mathbf{w} \text{ is a row in } T\}.$$

We can use tableaux to define mappings between relations over the same scheme. Consider a relation $\Phi_R$ over $S = R \cup \{f_{\phi_R}\} = \{A_1, A_2, \ldots, A_l, f_{\phi_R}\}$. Relation $\Phi_R$ is defined by a joint probability distribution $\phi_R$ on $R$. Let $\mathbf{w}_d = < a_1, a_2, \ldots, a_l >$ be the tuple of all distinguished variables. (The tuple $\mathbf{w}_d$ is not necessarily in $T$). Let $\{p_1, p_2, \ldots, p_k\}$ be the set of variables corresponding to the attribute $f_{\phi_R}$. Given $\Phi_R$, we can define a relation $T(\Phi_R)$ over $S$ as follows:

$$\begin{aligned}
T(\Phi_R) &= \{< \delta(a_1), \delta(a_2), \ldots, \delta(a_l), \psi_R(\delta(\mathbf{w}_d)) > \\
&\quad \mid \delta(T) \subseteq \Phi_R\}^{\downarrow R}, \quad (3)
\end{aligned}$$

where the values of the function $\psi_R(\delta(\mathbf{w}_d))$ may depend on $\delta(p_1), \delta(p_2), \ldots, \delta(p_k)$, where $\delta(p_i) = \phi_R(\delta(\mathbf{w}_i[R]))$.

It is always possible to find a tableau $T$ and an appropriate function $\psi$ for representing a marginalize-product-join mapping $m_\mathbf{R}$ defined by:

$$m_\mathbf{R}(\Phi_R) = (\ldots((\Phi_R^{\downarrow R_1} \otimes \Phi_R^{\downarrow R_2}) \otimes \Phi_R^{\downarrow R_3}) \ldots \otimes \Phi_R^{\downarrow R_N}),$$

where $\mathbf{R} = \{R_1, R_2, \ldots, R_N\}$ and $R = R_1 \cup R_2 \cup \ldots \cup R_N = \{A_1, A_2, \ldots, A_l\}$. The relations $\Phi_R^{\downarrow R_i}$ are marginals of relation $\Phi_R$ over $S = R \cup \{f_{\phi_R}\}$. Recall that $\Phi_R$ represents a decomposable joint probability distribution $\phi_R$. This means that the corresponding hypergraph $\mathbf{R}$ is a hypertree. In defining the mapping $m_\mathbf{R}$, we have tacitly assumed that the sequence, $R_1, R_2, \ldots, R_N$, is a tree construction ordering for the hypertree $\mathbf{R}$, and $\mathcal{L} = \{R_{j(2)} \cap R_2, R_{j(3)} \cap R_3, \ldots, R_{j(N)} \cap R_N\}$ is its intersection set.

The tableau for $m_\mathbf{R}$, $T_\mathbf{R}$, is defined as follows. The scheme for $T_\mathbf{R}$ is $S = R \cup \{f_{\phi_R}\}$. $T_\mathbf{R}$ has $N$ rows, $\mathbf{w}_1, \mathbf{w}_2, \ldots, \mathbf{w}_N$. Row $\mathbf{w}_i$ has the distinguished variable $a_j$ in the $A_j$-column exactly when $A_j \in R_i$. The

$$T_\mathbf{R} = \begin{array}{|cccc|l|} \hline
A_1 & A_2 & A_3 & A_4 & f_{\phi_R} \\ \hline
a_1 & a_2 & b_1 & b_2 & p_1 = \phi_R(a_1, a_2, b_1, b_2) \\
b_3 & a_2 & a_3 & b_4 & p_2 = \phi_R(b_3, a_2, a_3, b_4) \\
b_5 & b_6 & a_3 & a_4 & p_3 = \phi_R(b_5, b_6, a_3, a_4) \\ \hline
\end{array}$$

Figure 2: The tableau $T_\mathbf{R}$ for $m_\mathbf{R}$ with $\mathbf{R} = \{A_1A_2, A_2A_3, A_3A_4\}$.

variable of the attribute $f_{\phi_R}$ in row $\mathbf{w}_i$ is defined by $p_i = \phi_R(\mathbf{w}_i[R])$. The remaining nondistinguished variables in $\mathbf{w}_i$ are unique and they appear in no other rows of $T_\mathbf{R}$.

*Example 1.* Let $\mathbf{R} = \{R_1, R_2, R_3\} = \{A_1A_2, A_2A_3, A_3A_4\}$. The tableau $T_\mathbf{R}$ for $m_\mathbf{R}$ is shown in Figure 2. $\square$

To complete the definition for the mapping $T_\mathbf{R}$, we need to choose suitable values for the function $\psi_R$ in the individual rows, $< \delta(a_1), \delta(a_2), \delta(a_3), \delta(a_4), \psi_R(\delta(\mathbf{w}_d)) >$, such that $T_\mathbf{R}(\Phi_R) = m_\mathbf{R}(\Phi_R)$ for any relation $\Phi_R$ over scheme $S$. Recall that $\delta(p_i) = \phi_R(\delta(\mathbf{w}_i))$. For this purpose, we define $\psi_R(\delta(\mathbf{w}_d))$ as follows:

$$\begin{aligned}
&\psi_R(\delta(\mathbf{w}_d)) \\
&= \psi_R(\delta(a_1), \ldots, \delta(a_l)) \quad (4) \\
&= \frac{\phi_R(\delta(\mathbf{w}_1[R])) \ldots \phi_R(\delta(\mathbf{w}_N[R]))}{\phi_R(\delta(\mathbf{w}_d[R_{j(2)} \cap R_2])) \ldots \phi_R(\delta(\mathbf{w}_d[R_{j(N)} \cap R_N]))}.
\end{aligned}$$

Note that if $\delta(T_\mathbf{R}) \subseteq \Phi_R$, by substituting the values $\psi_R(\delta(\mathbf{w}_d[R]))$ defined by Equation 4 into Equation 3, it immediately follows that $T_\mathbf{R}$ satisfies the condition $T_\mathbf{R}(\Phi_R) = m_\mathbf{R}(\Phi_R)$. That is, the marginalize-product-join mapping $m_\mathbf{R}$ and the tableau $T_\mathbf{R}$ define the *same* function between relations over scheme $S$.

Note that the tableau $T_I$, containing only the row $\mathbf{w} = < \mathbf{w}_d, \psi_R(\mathbf{w}_d) >$ with $\psi_R(\mathbf{w}_d) = \phi_R(\mathbf{w}_d)$, is the *identity mapping* on all relations $\Phi_R$ over the same scheme.

### 3.3 THE CHASE

We now describe a computation method, the *chase*, for testing implication of dependencies (independencies). We will focus primarily on logical implications of GAJDs.

Let $\mathbf{P} = SAT(\mathbf{C})$ be the set of relations $\Phi_R$ defined by a set $\mathbf{C}$ of constraints. We say tableaux $T_1$ and $T_2$ are *equivalent* on $\mathbf{P}$, written $T_1 \equiv_\mathbf{P} T_2$, if $T_1(\Phi_R) = T_2(\Phi_R)$ for all $\Phi_R$ in $\mathbf{P}$.

We first consider methods for modifying tableaux while preserving equivalence. A *transformation rule* for **C** is a method for changing a tableau $T$ to a tableau $T'$ with $T \equiv_\mathbf{P} T'$. When **P** is the set of all relations, the set of all possible transformation rules is very limited. However, when the set of admissible relations is restricted, more rules are available. In this paper, we assume **C** is a set of GAJDs, and consider only *one* kind of transformation rules, the J-rules.

A *J-rule* corresponding to a GAJD, $\otimes \mathbf{Q}[\Phi_R]$, is defined as follows: Let the sequence $Q_1, Q_2, \ldots, Q_q$, be a tree construction ordering for the hypergraph $\mathbf{Q} = \{Q_1, Q_2, \ldots, Q_q\}$, and let $\mathcal{L}_Q = \{Q_{j(2)} \cap Q_2, Q_{j(3)} \cap Q_3, \ldots, Q_{j(q)} \cap Q_q\}$ be its intersection set. Consider a tableau $T$ over the scheme $S = Q_1 \cup Q_2 \cup \ldots Q_q \cup \{f_{\phi_Q}\} = Q \cup \{f_{\phi_Q}\} = R \cup \{f_{\phi_R}\} = \{A_1, A_2, \ldots, A_l, f_{\phi_R}\}$. Note that $Q = R$. The variable $p_i$ of the attribute $f_{\phi_R}$ for row $\mathbf{w_i}$ in $T$ is equal to $p_i = \phi_R(\mathbf{w}_i[R])$. Here we view tableau $T$ as a *relation* over $S$. We say that rows $\mathbf{w}_{k_1}, \mathbf{w}_{k_2}, \ldots, \mathbf{w}_{k_q}$ of $T$ (not necessarily distinct) are *joinable* on $\mathbf{Q}$ if there exists a row $\mathbf{w}$ not in $T$ that agrees with $\mathbf{w}_{k_i}$ on $Q_i$, i.e., $\mathbf{w}[Q_i] = \mathbf{w}_{k_i}[Q_i], 1 \le i \le q$. The variable $\phi_R(\mathbf{w}[R])$ in row $\mathbf{w}$ is defined by:

$$\phi_R(\mathbf{w}[R]) = \frac{\phi_R(\mathbf{w}_{k_1}[Q_1]) \ldots \phi_R(\mathbf{w}_{k_q}[Q_q])}{\phi_R(\mathbf{w}[Q_{j(2)} \cap Q_2]) \ldots \phi_R(\mathbf{w}[Q_{j(q)} \cap Q_q])}. \quad (5)$$

Add this row $\mathbf{w}$ to $T$ to form tableau $T'$.

Equation 5 can be equivalently expressed as:

$$\phi_R(\mathbf{w}[R]) \qquad (6)$$
$$= \frac{\phi_R^{\downarrow Q_1}(\mathbf{w}_{k_1}[Q]) \ldots \phi_R^{\downarrow Q_q}(\mathbf{w}_{k_q}[Q])}{\phi_R(\mathbf{w}[Q_{j(2)} \cap Q_2]) \ldots \phi_R(\mathbf{w}[Q_{j(q)} \cap Q_q])}.$$

It should be noted that $Q_{j(i)} \cap Q_i = \mathbf{w}_{j(i)} \cap \mathbf{w}_i$, $2 \le i \le q$.

*Example 2.* Consider the tableau $T_\mathbf{R}$ given in Figure 2. The J-rule for the GAJD, $\otimes\{A_1A_2, A_2A_3A_4\}[\Phi_R]$, can be applied to the first row $\mathbf{w}_1 = <a_1, a_2, b_1, b_2, \phi_R(a_1, a_2, b_1, b_2)>$ and the second row $\mathbf{w}_2 = <b_3, a_2, a_3, b_4, \phi_R(b_3, a_2, a_3, b_4)>$ of $T_\mathbf{R}$ to generate the row $\mathbf{w}_3 = <a_1, a_2, a_3, b_4, \phi_R(a_1, a_2, a_3, b_4)>$, where

$$\phi_R(\mathbf{w}_3[A_1A_2A_3A_4])$$
$$= \phi_R(a_1, a_2, a_3, b_4)$$
$$= \frac{\phi_R(\mathbf{w}_1[A_1A_2]) \cdot \phi_R(\mathbf{w}_2[A_2A_3A_4])}{\phi_R(\mathbf{w}_3[\{A_1A_2\} \cap \{A_2A_3A_4\}])}$$
$$= \frac{\phi_R(a_1, a_2) \cdot \phi_R(a_2, a_3, b_4)}{\phi_R(a_2)}$$
$$= \frac{\phi_R^{\downarrow A_1A_2}(a_1, a_2, b_1, b_2) \cdot \phi_R^{\downarrow A_2A_3A_4}(b_3, a_2, a_3, b_4)}{\phi_R(a_2)}$$

$$T_\mathbf{R}' = \begin{array}{|cccc|c|} \hline A_1 & A_2 & A_3 & A_4 & f_{\phi_R} \\ \hline a_1 & a_2 & b_1 & b_2 & \phi_R(a_1, a_2, b_1, b_2) \\ b_3 & a_2 & a_3 & b_4 & \phi_R(b_3, a_2, a_3, b_4) \\ b_5 & b_6 & a_3 & a_4 & \phi_R(b_5, b_6, a_3, a_4) \\ a_1 & a_2 & a_3 & b_4 & \phi_R(a_1, a_2, a_3, b_4) \\ \hline \end{array}$$

Figure 3: The result of applying the J-rule for the GAJD, $\otimes\{A_1A_2, A_2A_3A_4\}[\Phi_R]$, to the tableau $T_\mathbf{R}$ in Figure 2.

Tableau $T_\mathbf{R}'$ in Figure 3 is the result of this application. Note that we cannot construct the row $\mathbf{w} = <a_1, a_2, a_3, a_4, \phi_R(a_1, a_2, a_3, a_4)>$ since no J-rule exists which applies to attribute $A_3$. □

Clearly, when the set **P** of relations over $S$ is defined by a set **C** of GAJDs, i.e., $\mathbf{P} = SAT(\mathbf{C})$, the corresponding J-rules can be used to generate for each tableau another tableau. It can be shown that the J-rules associated with a set **C** of GAJDs are a *Finite Church-Rosser* (FCR) system [11]. That is, the resultant tableau $T^*$ is unique, independent of the order in which the rules were applied. The tableau $T^*$ called the *chase of $T$ under* **C**, written $chase_C(T)$, is obtained from $T$ by repeated applications of the rules in **C** until no new row is being generated.

Let $T_0, T_1, T_2, \ldots, T_n$ denote a *generating sequence* for $T$ in the *chase* such that $T_0 = T$, $T_i$ is obtained from $T_{i-1}$ by an application of a rule in **C**, and $T_n = T^*$. It is not difficult to see from Equation 6 that $T_{i-1} \equiv_\mathbf{P} T_i, 1 \le i \le n$. This means that $T \equiv_\mathbf{P} T^*$.

## 4 TESTING IMPLICATION OF DEPENDENCIES

In this section, we demonstrate that the *chase* is a remarkable tool for reasoning about dependencies. In particular, we show how it can be used for testing logical implications of probabilistic dependencies (independencies). It can also be used to derive nontrivial theoretical results.

We desire a means to test when all the relations $\Phi_R$ in **P** described by a set of constraints **C** (i.e., $\mathbf{P} = SAT(\mathbf{C})$), satisfy a particular GAJD, say $\otimes \mathbf{R}[\Phi_R]$. That is, we want to test if $\mathbf{C} \models \otimes \mathbf{R}[\Phi_R]$ or $T_\mathbf{R}(\Phi_R) = \Phi_R$ holds for all $\Phi_R$'s in **P**. When this holds, the tableau $T_\mathbf{R}$ for the GAJD, $\otimes \mathbf{R}[\Phi_R]$, is equivalent to the identity mapping on **P**. Testing for this condition amounts to showing whether or not $T_\mathbf{R}^* = chase_C(T_\mathbf{R})$ contains the row $<a_1, a_2, \ldots, a_l, \phi_R(\mathbf{w}_d)>$.

Suppose $T_\mathbf{R}^*$ does contain the row $\mathbf{w} = <\mathbf{w}_d, \phi_R(\mathbf{w}_d)>$, where $\mathbf{w}_d = <a_1, a_2, \ldots, a_l>$ and $\mid R \mid = l$. By assumption, **R** is a hypertree. Let $R_1, R_2, \ldots, R_N$ be a hypertree construction ordering for **R**, i.e., $R_i$ is a twig of $\{R_1, R_2, \ldots, R_i\}$, $2 \leq i \leq N$. Since the *chase* procedure is a FCR system, we may assume that the row **w** was obtained by applying $n \leq N-1$ distinct J-rules, $J_1, J_2, \ldots, J_n$, in **C** *sequentially* according to the ordering, $R_N, R_{N-1}, \ldots, R_1$. For convenience, we label the rows, $\mathbf{w}_N, \mathbf{w}_{N-1}, \ldots, \mathbf{w}_1$, of $T_R$ corresponding to this ordering. Assume that rule $J_1$ corresponds to the GAJD, say $\otimes \mathbf{Q}[\Phi_R]$, where $\mathbf{Q} = \{R_N, R_{N-1}, \ldots, R_{q+1}, Q_q\}$, $\mid \mathbf{Q} \mid = N-q+1$, and $R = R_N \cup R_{N-1} \cup \ldots \cup R_{q+1} \cup Q_q$. Applying rule $J_1$ to the rows, $\mathbf{w}_N, \mathbf{w}_{N-1}, \ldots, \mathbf{w}_{q+1}, \mathbf{w}_q$ of $T_R$, we obtain from Equation 5 the row $\mathbf{w}'_q = \ <\mathbf{w}'[R], \phi_R(\mathbf{w}'[R])>$, where

$$\mathbf{w}'_q[R] = \mathbf{w}_N[R_N] \cup \ldots \cup \mathbf{w}_{q+1}[R_{q+1}] \cup \mathbf{w}_q[Q_q],$$

and

$$\phi_R(\mathbf{w}'_q[R])$$
$$= \frac{\phi_R(\mathbf{w}_N[R_N]) \ldots \phi_R(\mathbf{w}_q[Q_q])}{\phi_R(\mathbf{w}'_q[R_{j(N)} \cap R_N]) \ldots \phi_R(\mathbf{w}'_q[R_{j(q+1)} \cap R_{q+1}])}.$$

Next, we apply rule $J_2$ for the GAJD, say $\otimes \mathbf{S}[\Phi_R]$, to the rows, $\mathbf{w}'_q, \mathbf{w}_{q-1}, \ldots, \mathbf{w}_s$, where $\mid \mathbf{S} \mid = q-s+1$, and so on. Finally we obtain:

$$\begin{aligned}&\phi_R(\mathbf{w}_d)\\=\ &\phi_R(a_1, a_2, \ldots, a_l) \qquad (7)\\=\ &\frac{\phi_R(\mathbf{w}_N[R_N]) \cdot \ldots \cdot \phi_R(\mathbf{w}_1[R_1])}{\phi_R(\mathbf{w}_d[R_{j(N)} \cap R_N]) \ldots \phi_R(\mathbf{w}_d[R_{j(2)} \cap R_2])},\end{aligned}$$

where $\mathbf{w}_N, \mathbf{w}_{N-1}, \ldots, \mathbf{w}_1$ are the *original* rows of $T_R$ corresponding to the relational schemes $R_N, R_{N-1}, \ldots, R_1$, respectively. Since, by the construction of $T_R$, $\mathbf{w}_i[R_i]$ contains the distinguished $a_k$ in the $A_k$-column exactly when $A_k \in R_i$, we have $\mathbf{w}_i[R_i] = \mathbf{w}_d[R_i]$, $1 \leq i \leq N$. Thus, Equation 7 can be written as:

$$\begin{aligned}&\phi_R(\mathbf{w}_d) \qquad\qquad (8)\\=\ &\frac{\phi_R(\mathbf{w}_d[R_N]) \ldots \phi_R(\mathbf{w}_d[R_1])}{\phi_R(\mathbf{w}_d[R_{j(N)} \cap R_N]) \ldots \phi_R(\mathbf{w}_d[R_{j(2)} \cap R_2])}.\end{aligned}$$

This means that $\phi_R$ is a decomposable probability distribution. Thus, $T_R^*$ is indeed the identity mapping on **P**. Since $T_R \equiv_\mathbf{P} T_R^*$, the condition $T_R(\Phi_R) = \Phi_R$ is satisfied by all the relations $\Phi_R$ in **P**. Similarly, we can show that the converse is true. That is, if **C** implies the GAJD, $\otimes \mathbf{R}[\Phi_R], T_R^*$ must contain the row $\mathbf{w} = \ <a_1, a_2, \ldots, a_l, \phi_R(\mathbf{w}_d)>$, where $\phi_R(\mathbf{w}_d)$ is defined by Equation 8.

It is important to note that the test for whether $T_R^*$ contains the row $<a_1, a_2, \ldots, a_l, \phi_R(\mathbf{w}_d)>$ can simply

$$T''_\mathbf{R} = \begin{array}{|c|c|c|c|c|}\hline A_1 & A_2 & A_3 & A_4 & f_{\phi_R}\\ \hline a_1 & a_2 & b_1 & b_2 & \phi_R(a_1, a_2, b_1, b_2)\\ b_3 & a_2 & a_3 & b_4 & \phi_R(b_3, a_2, a_3, b_4)\\ b_5 & b_6 & a_3 & a_4 & \phi_R(b_5, b_6, a_3, a_4)\\ a_1 & a_2 & a_3 & b_4 & \phi_R(a_1, a_2, a_3, b_4)\\ a_1 & a_2 & a_3 & a_4 & \phi_R(a_1, a_2, a_3, a_4)\\ \hline\end{array}$$

Figure 4: The result of applying the *J-rule* $\otimes \{A_1A_2A_3, A_3A_4\}$ to the third and fourth rows of tableau $T'_\mathbf{R}$ in Figure 3.

be done by checking whether there exists the row **w** in $T_R^*$ such that $\mathbf{w}[R] = <a_1, a_2, \ldots, a_l>$.

The above results are summarized in the following theorem.

**Theorem 2** *Let* **C** *and* $\{\otimes \mathbf{R}[\Phi_R]\}$ *be sets of GAJDs over the scheme* $S = R \cup \{f_{\phi_R}\}$. *Let* $T_\mathbf{R}$ *be the tableau corresponding to the GAJD,* $\otimes \mathbf{R}[\Phi_R]$, *and let* $T_\mathbf{R}^* = chase_C(T_\mathbf{R})$ *be the result of the chase, where* $\mathbf{R} = \{R_1, R_2, \ldots, R_N\}$ *and* $R = R_1 \cup R_2 \cup \ldots \cup R_N$. *Then* $\mathbf{C} \models \otimes \mathbf{R}[\Phi_R]$ *iff there exists a row* **w** *in* $T_\mathbf{R}^*$ *such that* $\mathbf{w}[R] = \mathbf{w}_d = <a_1, a_2, \ldots, a_l>$, *where* $\mid R \mid = d$.

*Example 3.*  Let $T_\mathbf{R}$ be the tableau corresponding to the GAJD, $\otimes \mathbf{R} = \otimes \{A_1A_2, A_2A_3, A_3A_4\}$, as shown in Figure 2, and let $\mathbf{C} = \{\otimes\{A_1A_2, A_2A_3A_4\}, \otimes\{A_1A_2A_3, A_3A_4\}\}$ be a set of constraints. (For simplicity, $\otimes \mathbf{R}[\Phi_R]$ is written as $\otimes \mathbf{R}$.) As in Example 2, we can apply the *J-rule* for $\otimes\{A_1A_2, A_2A_3A_4\}$ to the first and second rows of $T_\mathbf{R}$ to produce the row $\mathbf{w}_3 = <a_1, a_2, a_3, b_4, \phi_R(a_1, a_2, a_3, b_4)>$ in Figure 3. Similarly, the *J-rule* $\otimes\{A_1A_2A_3, A_3A_4\}$ can be applied to the third and fourth rows of $T'_\mathbf{R}$ in Figure 3 to generate the row $\mathbf{w}_4 = <a_1, a_2, a_3, a_4, \phi_R(a_1, a_2, a_3, a_4)>$. Tableau $T''_\mathbf{R}$ in Figure 4 is the result of this application.

Note that the expression $\phi_R(a_1, a_2, a_3, a_4)$ in Figure 4 can be expressed as:

$$\begin{aligned}&\phi_R(a_1, a_2, a_3, a_4)\\=\ &(\phi_R^{\downarrow A_1A_2}(a_1, a_2, b_1, b_2) \cdot \phi_R^{\downarrow A_2A_3A_4}(b_3, a_2, a_3, b_4))^{\downarrow A_1A_2A_3} \cdot\\&\phi_R^{\downarrow A_3A_4}(b_5, b_6, a_3, a_4)\ /\ \phi_R(a_2) \cdot \phi_R(a_3)\\=\ &\frac{\phi_R(a_1, a_2) \cdot \phi_R(a_2, a_3) \cdot \phi_R(a_3, a_4)}{\phi_R(a_2) \cdot \phi_R(a_3)}.\end{aligned}$$

Since $T''_\mathbf{R}$ contains the row $\mathbf{w} = <a_1, a_2, a_3, a_4, \phi_R(a_1, a_2, a_3, a_4)>$, by Theorem 2, one can therefore conclude that

$$\{\otimes\{A_1A_2, A_2A_3A_4\}, \otimes\{A_1A_2A_3, A_3A_4\}\}$$
$$\models \otimes\{A_1A_2, A_2A_3, A_3A_4\}. \quad \square$$

## 5  CONCLUSION

We have shown in this paper that the *chase* technique provides an alternative method for testing logical implication of probabilistic dependencies. Although the chase computation may need exponential time in the worst case, our preliminary investigations indicate that this technique is a powerful tool in the study of dependencies and optimization problems in probabilistic reasoning. More importantly, perhaps, the present study further demonstrates the close relationship between relational and probabilistic knowledge systems.

## References

[1] A.V. Aho, Y. Sagiv and J.D. Ullman, "Efficient optimization of a class of relational expressions," *ACM Transactions on Database Systems*, vol. 4, no. 4, pp. 435-454, 1979.

[2] C. Beeri, R. Fagin and J.H. Howard, "A complete axiomatization for functional and multivalued dependencies," Proc. ACM SIGMOD Symposium on Management of Data, pp. 47-61, 1977.

[3] C. Beeri, R. Fagin, D. Maier and M. Yannakakis, "On the desirability of acyclic database schemes," *Journal of the Association for Computing Machinery*, vol. 30, no. 3, pp. 479-513, 1983.

[4] C. Berge, *Graphs and Hypergraphs*. North Holland, 1973.

[5] G.F. Cooper, "The computational complexity of probabilistic inference using Bayesian belief networks," *Artificial Intelligence*, vol. 42, no. 2-3, pp. 393-402, 1990.

[6] C. Delobel, "Normalization and hierarchical dependencies in the relational data model," *ACM Transactions on Database Systems*, vol. 3, no. 3, pp. 201-222, 1978.

[7] R. Fagin, A.O. Mendelzon and J.D. Ullman, "A simplified universal relation assumption and its properties," *ACM Transactions on Database Systems*, vol. 7, no. 3, pp. 343-360, 1982.

[8] R. Fagin, "Multivalued dependencies and a new normal form for relational databases," *ACM Transactions on Database Systems*, vol. 2, no. 3, pp. 262-278, 1977.

[9] P. Hajek, T. Havranek and R. Jirousek, *Uncertain Information Processing in Expert Systems*. CRC Press, 1992.

[10] F.V. Jensen, "Junction trees – a new characterization of decomposable hypergraphs," Research Report, JUDEX, Aalborg, Denmark, 1988.

[11] D. Maier, *The Theory of Relational Databases*. Computer Science Press, 1983.

[12] R.E. Neapolitan, *Probabilistic Reasoning in Expert Systems*. John Wiley & sons, Inc., 1990.

[13] J. Pearl, "Fusion, propagation and structuring in belief networks," *Artificial Intelligence*, vol. 29, no. 3, pp. 241-288, 1986.

[14] J. Pearl, *Probabilistic Reasoning in Intelligent Systems*. Morgan Kaufmann, 1988.

[15] J. Pearl and A. Paz, "GRAPHOIDS: a graph-based logic for reasoning about relevance relations," Technical Report 850038 (R-53-L), Cognitive Systems Laboratory, UCLA, 1988.

[16] Y. Sagiv and F. Walecka, "Subset dependencies and a complete result for a subclass of embedded multivalued dependencies," *Journal of ACM*, vol. 20, no. 1, pp. 103-117, 1982.

[17] M. Studeny, *Kybernetika*, vol. 25, no. 1-3, pp. 72-79, 1990.

[18] G. Shafer, "An axiomatic study of computation in hypertrees," School of Business Working Paper Series, (No. 232), University of Kansas, Lawrence, 1991.

[19] S.K.M. Wong, Z.W. Wang, "On axiomatization of probabilistic conditional independence," Proc. Tenth Conference on Uncertainty in Artificial Intelligence, 591-597, 1994.

[20] S.K.M. Wong, Y. Xiang and X. Nie, "Representation of bayesian networks as relational databases," Proc. Fifth International Conference Information Processing and Management of Uncertainty in Knowledge-Based Systems, 159-165, 1994.

[21] S.K.M. Wong, C.J. Butz and Y. Xiang, "A method for implementing a probabilistic model as a relational database," Proc. Eleventh Conference on Uncertainty in Artificial Intelligence, 556-564, 1995.